\documentclass[1p]{elsarticle}

\usepackage{hyperref}
\usepackage{amsmath,amssymb, amsfonts,empheq,mathtools,mathrsfs,nccmath}
\usepackage{float,subfigure, wrapfig}  
\usepackage{arydshln,multirow, multicol, rotating, longtable,lscape,booktabs}  
\usepackage{relsize} 
\usepackage{color,xcolor} 
\usepackage{geometry} 
\usepackage[linesnumbered,boxed,ruled,commentsnumbered]{algorithm2e}
\usepackage{tikz} 
\usetikzlibrary{shapes.geometric, arrows} 
\usepackage{verbatim} 
\usepackage{bm}
\tikzset{
  treenode/.style = {shape=rectangle, rounded corners,
                     draw, align=center,
                     top color=white, bottom color=blue!20},
  root/.style     = {treenode, font=\Large, bottom color=red!30},
  env/.style      = {treenode, font=\ttfamily\normalsize},
  dummy/.style    = {circle,draw}
} 

\biboptions{authoryear}

\begin{document}
	
\begin{frontmatter}
	
\title{Parallel Hierarchical Transformer with Attention Alignment for Abstractive Multi-Document Summarization}

\author{Ye Ma \\
  \texttt{ye.ma@xjtlu.edu.cn} \\
  Lu Zong \\
  \texttt{lu.zong@xjtlu.edu.cn} \\}

\begin{abstract}
In comparison to single-document summarization, abstractive Multi-Document Summarization (MDS) brings challenges on the representation and coverage of its lengthy and linked sources. This study develops a Parallel Hierarchical Transformer (PHT) with attention alignment for MDS. By incorporating word- and paragraph-level multi-head attentions, the hierarchical architecture of PHT allows better processing of dependencies at both token and document levels. To guide the decoding towards a better coverage of the source documents, the attention-alignment mechanism is then introduced to calibrate beam search with predicted optimal attention distributions. Based on the WikiSum data, a comprehensive evaluation is conducted to test improvements on MDS by the proposed architecture. By better handling the inner- and cross-document information, results in both ROUGE and human evaluation suggest that our hierarchical model generates summaries of higher quality relative to other Transformer-based baselines at relatively low computational cost. 
\end{abstract}
\end{frontmatter}

\section{Introduction}
Since \citet{10.5555/2969033.2969173} propose the sequence-to-sequence (seq2seq) model for machine translation, the development of NLP applications has been almost inseparable from this framework. In the field of abstractive summarization, the seq2seq model is first applied by
\citet{rush-etal-2015-neural} to summarize sentences. With respect to the recent bloom of the attention mechanism and pre-trained models, great effort has been made to improve neural machine summarization upon extensions of seq2seq \citep{gehrmann-etal-2018-bottom, See2017,zhang-etal-2019-hibert}. With the promising results on single documents \citep{See2017, gehrmann-etal-2018-bottom, lewis-etal-2020-bart, PRADHAN2021218, LIAO2021228, LIANG2021128}, there are increasing recent attempts to study abstractive multi-document summarization (MDS) in the seq2seq framework \citep{liu2018generating,lebanoff-etal-2018-adapting,fabbri-etal-2019-multi,Liu_2019,ma2020multidocument}.

This study makes an exploratory attempt to improve the established abstractive summarization models for multi-document summarization (MDS) utilizing the Transformer \citep{vaswani2017attention} architecture. In comparison to single-document summarization, MDS brings challenges on the representation and coverage of its lengthy and linked sources. \citet{liu2018generating} propose a two-stage model to first extractively select the important paragraphs, then train the concatenated flat sequences using the Transformer-decoder with memory compressed attention (T-DMCA). Although the two-stage approach effectively reduces redundant information of source documents and retains salient information as inputs, it fails to take into account the cross-document relationship in its summaries. On the other hand, \citet{Liu_2019} propose a Hierarchical Transformer (HT) with local and global encoder layers to represent cross-token and cross-document information. Summaries are then generated based on a vanilla Transformer \citep{vaswani2017attention} by concatenating document-information-enriched token embeddings to a flat sequence. The essentially flat nature of the model leads to restrictions on learning dependencies of input sequences longer than 2000 tokens \citep{liu2018generating}. 

As a solution to better process the long-term dependency and cross-document relationship in MDS, this study develops a novel Parallel Hierarchical Transformer (PHT) with the paragraph-level attention alignment. Operationally, PHT first creates the word- and paragraph-level context vectors from a shared encoder, then generates summaries by the the word- and paragraph-level multi-head attentions parallel to each other in the decoder. In this way, PHT allows a pure hierarchical learning structure extended from the vanilla Transformer \citep{vaswani2017attention} to learn both cross-token and cross-document relationships. The word- and paragraph-level context vectors are then jointly used to generate target sequences in order to address the long-dependency problem of the flat structure, thus to permit extended length of input document. Our experiments show the sole PHT model has already the capacity to outperform other strong Transformer-based summarization models.


To address the coverage of the multi-document summaries, the decoding inference is further modified according to the proposed attention-alignment algorithm. As the original beam search prefers to generate typical and dull sentences to avoid making mistakes \citep{holtzman2019curious}, the paragraph-level attention alignment mechanism is designed to regulate generated summaries to attend to source contents with the optimal coverage of salient information. Inspired by Google's Neural Machine Translation (NMT) \citep{Wu2016Google}, attention alignment taps into the determination of the optimal attention distribution of source paragraphs on summaries, by predicting the reference attention from the source.  The score function of the beam search is then refined in order to select summaries closest to the predicted optimal attention distribution. With significantly elevated ROUGE scores, it is evident that incorporating the attention-alignment mechanism further enhances the quality of generated summaries with minor computational cost-added from a shallow attention-prediction model, of which inputs and labels are both extracted from the PHT model. 


With regards to the core target of developing an enhanced paradigm for multi-document summarization based on Transformer, the contribution of this study is twofold. First, the hierarchical architecture with parallel multi-head attentions is designed to represent and exchange token- and document-level information for the generation of summaries based on the lengthy inputs. The effectiveness of the PHT model is investigated relative to a variety of summarization models, in terms of the ability to capture cross-document relationship, computational efficiency and improvements on the summarization quality. Second, the paragraph-level attention-alignment mechanism is proposed to guide the generated summaries in the decoding stage to calibrate the original beam search according to the learned attention distribution. The merits of attention alignment are not only reflected by promoting the optimal coverage of the generated summary to the source, but also its practical value of low computational cost and potential to be adopted by other attention-based summarization models. 

The remaining of the paper is organized as follows. Section \ref{related} discusses the related work. Section \ref{parallel} and \ref{AAI} introduce the methodology associated with the Parallel Hierarchical Transformer and the attention-alignment mechanism. Section \ref{experiment} and \ref{results} describe the experimental setups and analyze the results. Section \ref{conclusion} concludes.

\section{Related Work}\label{related}

In general, hierarchical models are designed with strengthened capacity to handle lengthy inputs, which have been widely used in document classification \citep{yang-etal-2016-hierarchical} or large-document summariztion \citep{li-etal-2018-improving-neural} tasks. In the filed of MDS, hierarchical structures allow not only to represent massive source inputs, but also to capture cross-document relationships. \citet{fabbri-etal-2019-multi} use a hierarchical RNN structure with Maximal Marginal Relevance \citep{Carbonell:1998:UMD:290941.291025} to better select salient paragraphs and reduce repetitions in the summary. \citet{zhang-etal-2019-hibert} pre-train a hierarchical BERT \citep{devlin2018bert} by masking a sentence and using other sentences to generate the masked one. \citet{Liu_2019} propose a Hierarchical Transformer for multi-document summarization to enrich token embeddings with cross-document relationships. A vanilla Transformer \citep{vaswani2017attention} is then used to conduct summarization after combining the enriched token embeddings in a flat sequence \citep{liu2018generating}.

With the aim to improve the quality of seq2seq summaries, existing studies tend to focus on the coverage of salient contents. \citet{liu2018generating} use a text-ranking mechanism to extract important paragraphs, which are later input to the neural abstractive model. \citet{gehrmann-etal-2018-bottom} train a selector to predict the phrases ought to appear in the final summaries and use them as summarization inputs. \citet{Chen_2018} select and compress salient sentences that are later re-organized in the summaries. In addition to the two-stage extraction-abstraction approaches, attempts are made to build hybrid summarization models by incorporating the sentence-level attention \citep{Mei2015What,Cohan_2018,you2019improving}, graph neural networks \citep{tan-etal-2017-abstractive, LIANG2021128,li-etal-2020-leveraging-graph}, Maximal Marginal Relevance \citep{lebanoff-etal-2018-adapting,fabbri-etal-2019-multi} or reinforcement learning \citep{YAO201852}. Additionally, \citet{li-etal-2018-guiding} add representations of key words to the summarizaiton model. \citet{Hua_2019} use an abstractive summarization model to concatenate extracted key phrases. Moreover, some studies suggest to improve the summarizaiton quality by modifying the objective function to encourage salient words \citep{Pasunuru_2018} or to penalize repetitive generations \citep{See2017, welleck2019neural}. More details on representative methods of this sort and their differences with attention alignment are discussed in Section \ref{learnNMT}.

\section{Parallel Hierarchical Transformer}\label{parallel}

\begin{figure*}
\centering
\includegraphics[scale=0.4]{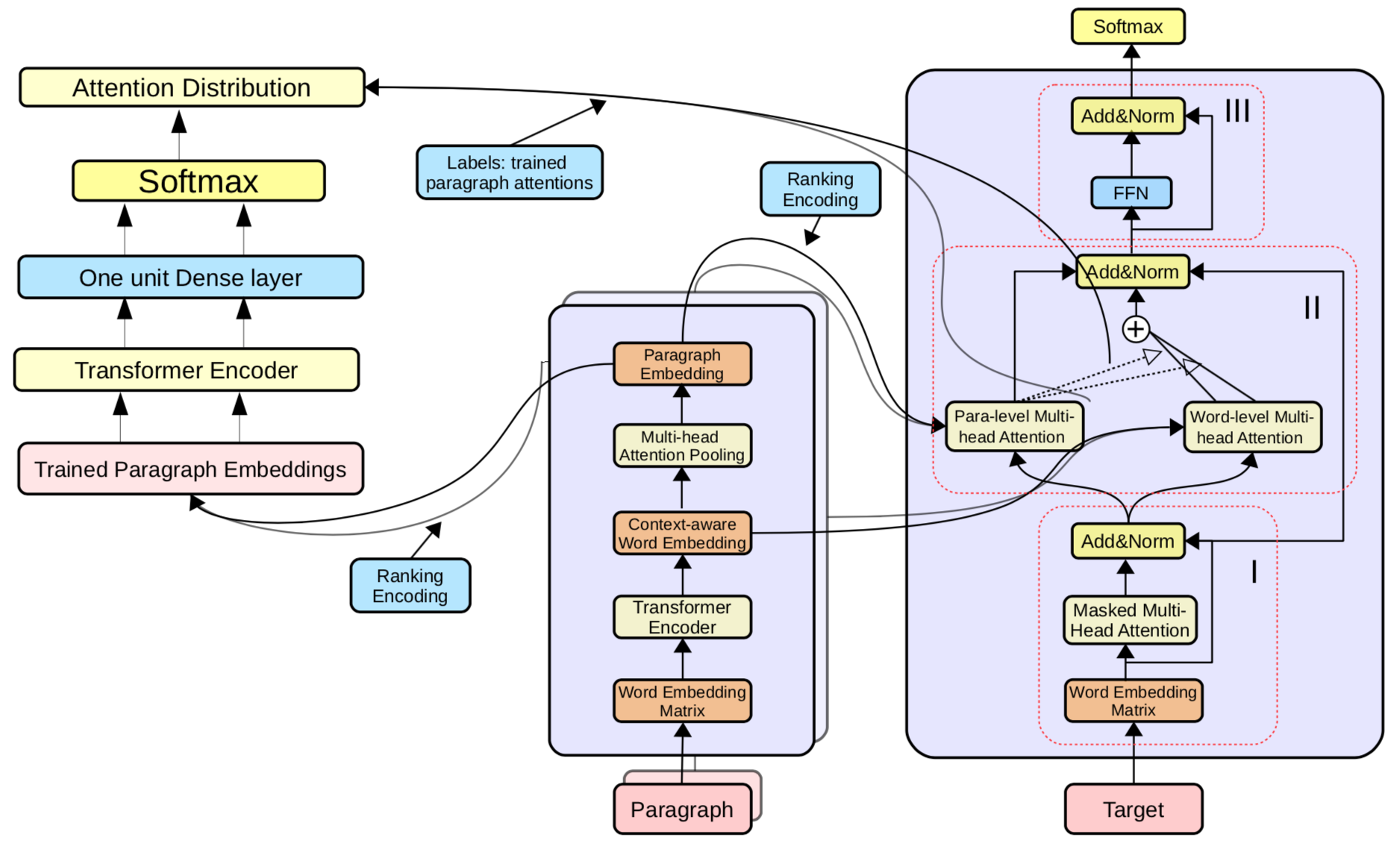}
\caption{Model flowchart with two input paragraphs. Middle and right: PHT encoder and decoder (See Section \ref{parallel}). Left: prediction model of the optimal attention distribution (See Section \ref{AAI}).}
\label{comb}
\end{figure*}

This section discusses the design of the Parallel Hierarchical Transformer for MDS. Figure \ref{comb} graphically presents the architecture of the proposed PHT. The encoder and decoder are displayed the second and third blocks on the left (highlighted in purple), respectively. The process of generating summaries is described as follows.

\subsection{Encoder}
As shown in Figure \ref{comb}, the PHT encoder is shared by all paragraphs and consist of two major units, i.e. the Transformer encoder and the additional Multi-head Attention Pooling layer, to obtain the token- and paragraph-embeddings, respectively. To be specific, context-aware word embeddings $\hm{C}_p \in \mathbb{R}^{n \times d} $ in the paragraph $p$ of length $n$ are first produced as the output of the Transformer encoder $\mathrm{TransE(\cdot)}$ based on the summation of word embeddings $\hm{W}_p \in \mathbb{R}^{n \times d}$ in the paragraph and their fixed positional encodings $\hm{E} \in \mathbb{R}^{n \times d}$ \citep{vaswani2017attention}.  
\begin{equation}
    \hm{C}_{p} = \mathrm{TransE}(\hm{W}_{p} + \hm{E})
\end{equation}
 The context-aware word embedding is then used to compute paragraph embeddings as well as being a part of inputs to the PHT decoder to calculate word-level cross attention. 

At the second step, PHT achieves paragraph embeddings based on $\hm{C}_{p}$ using {\bf Multi-head Attention Pooling}: 
\begin{equation}
    \hm{head}_{p}^h = \mathrm{HeadSplit}\left(\hm{C}_{p} \hm{W}_1 \right)
\end{equation}
\begin{equation}
    \hm{\phi}_p^h = \left(\mathrm{Softmax}( \hm{head}_p^h \hm{W}_2)\right)^\top \hm{head}_{p}^h
\end{equation}
\begin{equation}
    \hm{\phi}_p = \hm{W}_3 \left( \overset{heads}{\underset{h=1}{\Vert}} \hm{\phi}_p^h \right)
\end{equation}
\begin{equation}
    \hm{\phi}_p := \mathrm{LayerNorm}\left(\hm{\phi}_p + \mathrm{FFN}(\hm{\phi}_p)\right)
\end{equation}
where $\Vert$ represents concatenation, and $\hm{W}_1 \in \mathbb{R}^{d \times d}$, $\hm{W}_2 \in \mathbb{R}^{d_{head} \times 1}$, $\hm{W}_3 \in \mathbb{R}^{d \times d}$ are linear transformation parameters. Besides, $\hm{head}_{p}^h \in \mathbb{R}^{n \times d_{head}}$ denotes the $h_{th}$ attention head, and $\hm{\phi}_p^h \in \mathbb{R}^{d_{head}}$ is paragraph embedding of the head. These head embeddings are concatenated and fed to a two-layer feed forward network (FFN) with Relu activation function after linear transformation. The paragraph embedding is another input to the decoder for obtaining paragraph-level cross attention, together with the context-aware word embedding.

\subsection{Decoder}
The PHT decoder accepts three classes of inputs, namely the target summary, context-aware word embeddings in the $p_{th}$ paragraph $\hm{C}_p \in \mathbb{R}^{n \times d}$ where n is the length of the paragraph, and paragraph embeddings $\hm{\Phi} \in \mathbb{R}^{m \times d}$ where m is the number of paragraphs. Let $\hm{X}^{(\mathrm{I})} \in \mathbb{R}^{k \times d}$ denote the output of decoder part $\mathrm{I}$, where k is the length of target sequence or the number of time steps. Note that both the word embedding and vocabulary in the decoder part are shared with the encoder.

Different from the token-level ranking encoding \citep{Liu_2019}, we intend to incorporate the  information of paragraph importance to their embeddings. Specifically, ranking encoding\footnote{We directly use the ranked paragraphs provided by \citep{Liu_2019}.} $\hm{R} \in \mathbb{R}^{m \times d}$ created by the positional encoding function \citep{vaswani2017attention} are added to the original paragraph embeddings:
\begin{equation}
    \hm{\Phi} := \hm{\Phi} + \hm{R}
    \label{pb}
\end{equation}
PHT decoder consists of three parts. Similar to a vanilla Transformer \citep{vaswani2017attention}, the first and last parts of the PHT decoder are the masked multi-head attention and the feed forward network, whereas the second part includes two parallel-computing cross attention models to respectively capture the mutual information the target summary shares with source paragraphs and source words.

\textbf{Paragraph-level Cross Attention}. This cross attention model is to calculate the attention distribution that the decoder assigns to the paragraphs at each step, and at the same time represents the cross-paragraph relationships as paragraph-level context vectors. The query is the output of part $\mathrm{I}$: $\hm{X}^{(\mathrm{I})} \in \mathbb{R}^{k \times d}$ where $k$ is the length of the target sequence. The key and value are context-aware paragraph embeddings $\hm{\Phi}$. 
\begin{equation}
    \hm{X}^{\left<\mathrm{para}\right>}, \hm{A}^{\left<\mathrm{para}\right>} = \mathrm{MultiHead}\left(\hm{X}^{(\mathrm{I})}, \hm{\Phi}, \hm{\Phi}\right)
    \label{pa}
\end{equation}
where $\hm{X}^{\left<\mathrm{para}\right>} \in \mathbb{R}^{k \times d}$ is the weighted summation of paragraph embeddings, and $\hm{A}^{\left<\mathrm{para}\right>} \in \mathbb{R}^{k \times m}$ is the paragraphs attention weights \footnote{In this paper, average pooling is adopted to obtain the final attention from multi-head attention.}. 

\textbf{Word-level Cross Attention}. This cross attention mechanism aims at modeling how the decoder attends to source tokens of the paragraph. It could be considered as the local cross attention of each paragraph since their calculations of different paragraphs are independent. By comparison, paragraph-level cross attention refers to the global cross attention which captures the dependencies among paragraphs. Since the calculations of the two cross attention are based on different encoder outputs, they are non-interfering and parallel. Finally, the mechanism produces the word-level context vectors for each paragraphs. The query of the self attention is $\hm{X}^{(\mathrm{I})}$, whilst the key and value are context-aware word embeddings $\hm{C}_p$. 
\begin{equation}
    \hm{X}_p^{\left<\mathrm{word}\right>} = \mathrm{MultiHead}\left(\hm{X}^{(\mathrm{I})}, \hm{C}_p, \hm{C}_p\right)
\end{equation}
where $\hm{X}_p^{\left<\mathrm{word}\right>} \in \mathbb{R}^{k \times d}$ denotes the word-level context vectors of all time steps in the $p_{th}$ paragraph. 

\textbf{Multi-level Attention Fusion}.
Since the word-level cross attention model need to be implemented for each paragraph independently, there are totally $m$ word-level context vectors $\hm{X}_p^{\left<\mathrm{word}\right>}$, equivalently denoted as $\hm{\mathcal{X}}^{\left<\mathrm{word}\right>}  \in \mathbb{R}^{k \times d \times m}$. To fuse it with the paragraph-level context vectors $\hm{X}^{\left<\mathrm{para}\right>} \in \mathbb{R}^{k \times d}$ and part $\mathrm{I}$ hidden states $\hm{X}^{(\mathrm{I})} \in \mathbb{R}^{k \times d}$, we need to integrate $m$ groups of context vectors $\hm{\mathcal{X}}^{\left<\mathrm{word}\right>}$ to one group. The straightforward way is to use mean pooling or max pooling, but both may cause loss of context information. An alternative approach is the adaptive attention pooling but not conductive to the computational efficiency. To handle the two problems, we directly integrate  $\hm{\mathcal{X}}^{\left<\mathrm{word}\right>} $ with knowledge learned by the paragraph-level cross attention model, i.e., using paragraph attention $\hm{A}^{\left<\mathrm{para}\right>}$ to weight the context vectors $\hm{X}_p^{\left<\mathrm{word}\right>}$ of the corresponding paragraph. The related matrix calculation process is as follows:
\begin{equation}
    \hm{X}^{\left<\mathrm{int}\right>} = \hm{\mathcal{X}}^{\left<\mathrm{word}\right>} \hm{\mathcal{A}}^{\left<\mathrm{para}\right>},
\end{equation}
where $\hm{\mathcal{X}}^{\left<\mathrm{word}\right>}  \in \mathbb{R}^{k \times d \times m}$, $\hm{\mathcal{A}}^{\left<\mathrm{para}\right>} \in \mathbb{R}^{k \times m \times 1}$, and matrices are multiplied in the last two dimensions. The output of part $\mathrm{II}$ $\hm{X}^{(\mathrm{II})}$ is expressed as:
\begin{equation}
    \hm{X}^{(\mathrm{II})} = \mathrm{LayerNorm}\left(\hm{X}^{(\mathrm{I})} + \hm{X}^{\left<\mathrm{para}\right>} + \hm{X}^{\left<\mathrm{int}\right>}\right).
\end{equation}
With the outputs of part $\mathrm{II}$, we are able to proceed to part $\mathrm{III}$ and compute the final probability distributions.

\section{Attention-Alignment Mechanism}\label{AAI}
To further enhance the coverage of multi-document summarization, this section introduces the attention-alignment mechanism to guide the text decoding. The algorithm first predicts the optimal attention distribution of source paragraphs, then regulates the beam search according to the scoring function derived from the predicted attention distribution. Note that the attention-alignment mechanism is implemented after the training of PHT, in order to allow the extraction of the attention distribution from the trained parameters.

\subsection{Learn from Neural Machine Translation (NMT)}\label{learnNMT}
The idea of the Attention-Alignment is inspired by Google's NMT \citep{Wu2016Google}, where candidates in the beam search are re-ranked according to a refined score function with the length normalization
and coverage penalty. The penalty function is based on the assumption of one-to-one alignment in the translation so that $\sum_{t=1}^{T} \alpha_{t,i} = 1$, where $\alpha_{t,i}$ indicates the attention weight of the $t_{th}$ translated word on the $i_{th}$ source word. To penalize the situation that source words are not fully covered, i.e. the sum of attention weights is less than one, the coverage penalty is defined as:
\begin{equation}
    cp = \sum_{i=1}^{n} \log \left(\min\left(\sum_{t=1}^{T} \alpha_{t,i}, 1\right)\right)
    \label{gnmt}
\end{equation}

This assumption is not tenable for summarization as uniform coverage is no longer required. Pointer-generator \citep{See2017} re-defines the coverage loss for summarization as:
\begin{equation}
    cp_t = \sum_i \min\left(\alpha_{t,i}, \sum_{t'<t}\alpha_{t',i}\right)
\end{equation}
where $\alpha_{t,i}$ is the word-level attention distribution and $\sum_{t'<t}\alpha_{t',i}$ is the coverage vector. In this way, repeated attention is penalized according to the overlap between the attention distribution and the coverage til time step $t$.

\citet{li-etal-2018-improving-neural} further corporate this concept to their structural-coverage regularization, forcing the generation to focus on different source sentences to raise the diversity of the summary. In detail, the structural-coverage is defined as:
\begin{equation}
    strCov(\alpha_t) = 1 - \sum_i \min\left(\alpha_{t,i}, \sum_{t'<t}\alpha_{t',i}\right)
    \label{str_cov}
\end{equation}
which is rather similar to the coverage function of Pointer-generator \citep{See2017} except that \citet{li-etal-2018-improving-neural} consider the sentence-level attention $\alpha_{t,i}$ . 


In summary, both the Pointer-generator \citep{See2017} and structural-coverage regularization \citep{li-etal-2018-improving-neural} build their models based on the principle of searching for words/sentences that have previously attracted less attention to avoid repetition, thus to increase coverage. Comparing NMT's coverage penalty with the coverage functions in the aforementioned models \citep{See2017,li-etal-2018-improving-neural}, the restriction of summarization is rooted in the absence of the optimal attention distribution of contents, that maps a holistic layout of the summary with comprehensive coverage. This motivates us to develop the attention-alignment inference to address this matter.

\subsection{Paragraph-level Attention Alignment}
\subsubsection{Optimal attention distribution}
To explicitly express the coverage of source content, the first step of attention alignment is to use the encoded paragraph embeddings to predict the optimal attention distribution of input paragraphs. Specifically, the attention prediction model is trained from the label of paragraph-level attention distribution $\hm{\eta} \in  \mathbb{R}^{m}$ ($m$ is the number of paragraphs) calculated from paragraph attention weights $\hm{A}^{\left<\mathrm{para}\right>}$ in Eq.~\ref{pa} and
\begin{equation}
    \hm{A}^{\left<\mathrm{para}\right>} = \begin{bmatrix}
    \alpha_{1,1} & \dots  &\alpha_{1,m} \\
& \ddots & \vdots \\
\alpha_{k, 1} & & \alpha_{k,m}
\end{bmatrix} 
\end{equation}
 where $\alpha_{t,p} \in  \hm{A}^{\left<\mathrm{para}\right>}$ denotes the attention weight of the $t_{th}$ summary word on the $p_{th}$ source paragraph\footnote{In the case of multiple decoder layers, the final paragraph attention are the summation of paragraph attentions in each layer.}. 
 
\begin{equation}
    \eta_p = \frac{\sum_{t=1}^{k} \alpha_{t,p}}{\sum_{p=1}^m \sum_{t=1}^k \alpha_{t,p}}
    \label{goldd}
\end{equation}
\begin{equation}
    \hm{\eta} = [\eta_1,\cdots,\eta_p,\cdots,\eta_m]
\end{equation}
 Since the reference summary is known for training data, $\hm{\eta}$ is regarded as the optimal attention distribution and serves as the training label of the attention-prediction model. In other words, the labelling process only utilizes paragraph attention weights from the already-trained PHT parameters. Besides, the inputs of the attention-prediction model are extracted from the PHT, which are paragraph embeddings $\hm{\Phi}$ in Eq.~\ref{pb}. The training process is displayed in Figure \ref{comb}.

As for the construction of the attention prediction model, paragraph embeddings are first input to a Transformer-encoder to obtain the context-aware paragraph embeddings in order to make full usage of the context information between paragraphs. The context-aware paragraph embeddings are then linearly transformed and converted to $m$ (i.e. the number of paragraphs) units before normalized by softmax. Given the nature of the prediction is regression, mean square error (MSE) is used as the loss function.

During inference, the source paragraphs are first fed to the PHT encoder to obtain the paragraph embeddings $\hm{\Phi}$, based on which the trained attention-prediction model predicts the optimal attention distribution $\widehat{\hm{\eta}}$.

\subsubsection{Attention alignment score}

In line with NMT, the score function of the pure beam-search is modified taking into account the predicted optimal attention distribution. With length normalization and the attention-alignment score, the score of each candidate hypothesis is given by:
\begin{equation}
    score(\hm{y}) = \frac{\log\left(P(\hm{y}|\hm{x})\right)}{|\hm{y}|} + \beta * attAlign(\hm{y})
    \label{eq14}
\end{equation}
\begin{equation}
     attAlign(\hm{y}) = \sum_{p=1}^m \log\left(\min\left(\eta_p^{(\hm{y})}, \widehat{\eta}_p\right)\right)
\end{equation}
where $\hm{x}$ denotes the source, $\hm{y}$ refers to a candidate hypothesis, and $\widehat{\eta}_p \in \widehat{\hm{\eta}}$. Notably, different from the $\eta_p$ of reference, $\eta_p^{(\hm{y})}$ is obtained from the generated candidate summary.

This predicted optimal paragraph attention $\widehat{\eta}_p$ is compared with the paragraph attention $\eta_p^{(\hm{y})}$ in the real-time generation. Given any deviation between the real and optimal attention distributions, 
paragraphs that are assigned with underestimated attention place negative impacts on the overall scoring, whereas those with overestimated attention receive a constant score of $\widehat{\eta}_p$. Regarding the length normalization, we direct use the length $|\hm{y}|$, rather than $length^a$ \citep{Wu2016Google}, as it is empirically proven to be more suitable for longer summaries.

\subsubsection{Why using trained paragraph attention to form the optimal attention distribution?}\label{why}

An alternative way to obtain the optimal attention distribution is to use the paragraph ranking generated by an extractive model that predicts the probability each source sentence/paragraph appears in the final summary. However, the prediction of extractive probabilities is a separate unit from the summarization model which results in problematic inconsistency with the paragraph attention during the decoding process. 

\begin{figure}[htbp]
\centering
\includegraphics[width=8cm]{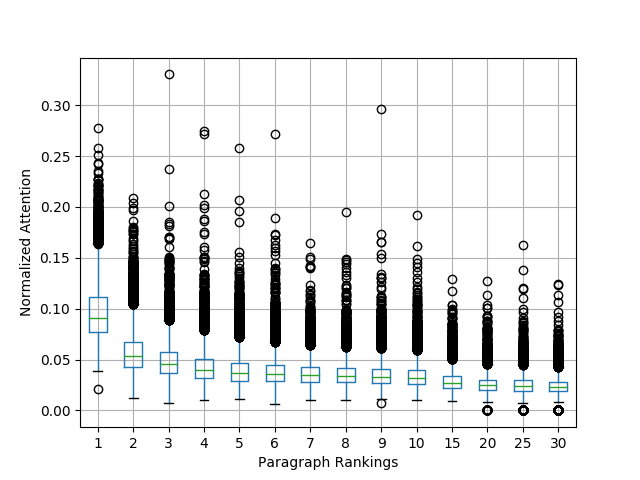}
\caption{Box plot of paragraph attention with different initial rankings.}
\label{box}
\end{figure}

To support this argument with empirical evidence, Figure \ref{box} randomly selects 10,000 training samples to compare the paragraph rankings \citep{Liu_2019} with the corresponding normalized paragraph attention by the trained HT decoder. In general, the decoder assigns higher attention to paragraphs with higher rankings. However, the outliers suggest that there are several cases of different judgements by the two approaches, which lead to potential conflicts during the inference given the inconsistent measures between the optimal and real attention. Therefore, we make our prediction model to learn paragraph attention from the trained decoder directly. These attentions are considered optimal as the training targets are gold summaries. The prediction model maps the connection between source documents and optimal attention distributions, to allow the predicted attention distribution to maximally approach the optimal attention distribution if the target is unknown.

In addition to the inconsistency problem, it is easier to acquire the attention weight since attention exists in almost all neural abstractive summarization models, among which many (especially single-document summarization) do not require extractive probabilities. Besides, the attention-prediction model directly extracts input vectors and labels from the summarization model, whereas the extractive method requires extra effort on representing inputs and making labels.

\section{Experiment Setup}\label{experiment}

\subsection{WikiSum Dataset}
Data sparsity has been the bottleneck of neural MDS models til WikiSum \citep{liu2018generating} came along. In this study, we use the ranked version of WikiSum provided by \cite{Liu_2019}. Each sample contains a short title, $40$ ranked paragraphs with a maximum length of $100$ tokens as source inputs, and a target summary with an average length of $140$ tokens. Consistent with \cite{Liu_2019}, the dataset is split with $1,579,360$ samples for training, $38,144$ for validation and $38,205$ for test. Subword tokenization \citep{bojanowski-etal-2017-enriching} is adopted to tokenize our vocabulary to $32,000$ subword units to better represent unseen words.

\subsection{Configuration}
The proposed PHT is trained on a single {\it 2080ti} with $0.3$ dropout rate and an Adam optimizer of 16,000 warm-up steps. We stack $3$-layer encoder-decoder of the PHT with $256$ hidden units, $1024$ FFN units and $4$ headers, top $3000$ tokens ($30$ paragraphs, $100$ tokens) are used to train the PHT for approximately $600,000$ steps. Checkpoints are saved every $20,000$ steps and the best result on the validation set is used to generate the final summary. All parameters are randomly initialized including token embeddings. In the decoding process, we take $3000$ tokens as inputs of PHT to generate summaries. Since the fixed positional encodings are used, so the attention-prediction model can accept inputs of dynamic length. We set the beam size to $5$ and terminate the inference til the length exceeds $200$. In addition, we disallow the repetition of trigrams, at the same time block two tokens (except commas) before the current step to prevent degeneration situations. For the attention prediction model, we construct a two-layer Transformer encoder with dropout rate $0.5$. The complete set of the training data is used to train the attention prediction for approximately $100,000$ steps. 

\subsection{Baselines}
- \textbf{Lead} \citep{nenkova2011automatic} is an extractive model that extracts the top $K$ tokens from the concatenated sequence. In MDS, we combine paragraphs in order and place the title at the beginning of the concatenated sequence.

- \textbf{LexRank} \citep{lexrank2004} is a widely-used graph-based extractive summarizer.

- \textbf{Flat Transformer (FT)} is the vanilla Transformer encoder-decoder model \citep{vaswani2017attention}. We adopt a $3$-layer Transformer in this study.

- \textbf{T-DMCA} \citep{liu2018generating} is a Transformer-decoder model that splits a concatenated sequence into segments, and uses a Memory Compressed Attention to exchange information among them. 

- \textbf{Transformer-XL} \citep{Dai_2019} is a language model that excels in handling excessively long sequences. This model improves the vanilla Transformer-decoder with the recurrent mechanism and relative positional encoding.

- \textbf{Liu's Hierarchical Transformer (Liu's HT)} \citep{Liu_2019} uses a hierarchical structure to enrich tokens with information from other paragraphs before inputting to the Flat Transformer. 

- \textbf{GraphSum} \citep{li-etal-2020-leveraging-graph} is an graph-based hierarchical transformer, where graph neural network is used to capture cross-document relationships.

- \textbf{Parallel Hierarchical Transformer (PHT)} is the model proposed in this paper. Different from Liu's HT, our hierarchical structure could compute token-level and paragraph-level dependencies, thus not requiring 


\subsubsection*{\underline{Decoding strategy}}

- \textit{PHT with vanilla beam search.} Beam search is a well-established baseline that is popularly-used in text generation tasks of all sorts.

- \textit{PHT with the structural-coverage (strCov).} As the original study \citep{li-etal-2018-improving-neural} summarizes single document using a hierarchical decoding algorithm to first decode sentence by sentence then realize the sentence word by word, we need to modify the regularization to adjust to the word-by-word inference. Therefore, we re-define $\alpha_{t,i}$ (in Eq.~\ref{str_cov}) from the attention of the $t_{th}$ generated sentence on the $i_{th}$ source sentence to the $t_{th}$ generated word on the $i_{th}$ source paragraph. To obtain an independent observation on the effect of the structural coverage, we skip the structural-compression regularization and other modifications on the loss function as discussed in \cite{li-etal-2018-improving-neural}.

- \textit{PHT with extractive probability (extProb)} also adopts attention alignment mechanism but replaces the learned optimal attention distribution $\widehat{\hm{\eta}}$ by the normalized extractive probabilities of paragraphs. We use the extractive method in Liu's HT to calculate these probabilities.

- \textit{PHT with the attention alignment mechanism (attAlign)} is PHT combined with the proposed attention-alignment mechanism. We probe the optimal value of the attention-alignment coefficient $\beta$ in Eq.~\ref{eq14} by a numerical comparison for $\beta \in [0.2:0.2:1]$ on ROUGE. The result of development set suggests the optimal value of $\beta$ is approximately 0.8 for the Wikisum dataset.

\section{Results}\label{results}
\subsection{Automatic Evaluation}
In this section, we adopt a group of widely-used evaluation metrics ROUGE \citep{lin-2004-rouge} to evaluate the MDS models. ROUGE-1 \& -2 and ROUGE-L $F_1$ scores are reported in Table \ref{tab:rou} assessing the informativeness and fluency of the summaries, respectively.

\begin{table}[htbp]
   \centering
    \begin{tabular}{lccc}
    \toprule
      Model & ROUGE-1 & ROUGE-2 & ROUGE-L  \\
      \midrule
        Lead & 36.40 & 16.66 & 32.95\\
        \midrule
        FT  & 40.30 & 18.67 & 32.84 \\
        T-DMCA & 41.09 & 19.78 & 33.31 \\
        Transformer-XL & 41.11 & 19.81 & 33.72 \\
        \midrule
        Liu's HT  & 40.83 & 19.41 & 33.26 \\
        1-layer PHT  &  41.02 & 19.82 & 33.28 \\
        PHT  & \textbf{41.99} & \textbf{20.44} & \textbf{34.50}\\
        \bottomrule
    \end{tabular}
   \caption{Average ROUGE $F_1$ scores of different summarization models.}
    \label{tab:rou}
\end{table}

As shown in Table \ref{tab:rou}, the extractive model Lead exhibits overall inferior performance in comparison to the abstractive models, except that it produces a 0.11-higher ROUGE-L than the Flat Transformer. Although Liu's HT improves FT with a hierarchical structure, it fails to outperform the two extended flat models, i.e. T-DMCA and Transformer-XL, that are developed to learn lengthier inputs. Moreover, T-DMCA and Transformer-XL report comparable results in terms of the informativeness (ROUGE-1 \& -2), whilst the latter outperforms the former by 0.41 in terms of the fluency (ROUGE-L).

Further, the proposed PHT model shows promising ROUGE results. Benefited from the pure hierarchical structure that allows prolonged token inputs, PHT outperforms Liu's HT in all domains of the ROUGE test. Moreover, the models' potential to be deepened is suggested by enhanced results of the 3-layer architecture over the 1-layer architecture. The ultimate 3-layer PHT stably surpasses T-DMCA and Transformer-XL, that are also tailored to handle long input sequences of 3,000 tokens, due to its hierarchical processing of token and document-level information.

\subsubsection{Comparing the decoding strategies}

    

\begin{table*}[htbp]
   \centering
    \begin{tabular}{lccc}
    \toprule
     { Parallel HT} & ROUGE-1 & ROUGE-2 & ROUGE-L  \\
       \midrule
     
        { + vanilla beam search}  & 41.99 & 20.44 & 34.50\\
        { + strCov \citep{li-etal-2018-improving-neural}} & 41.74 & 20.25 & 33.88 \\
         { + extProb} & 42.17 & 20.46 & 34.79 \\
        { + attAlign}  & \textbf{42.58} & \textbf{20.84} & \textbf{35.66}\\
        \bottomrule
    \end{tabular}
     \caption{Average ROUGE $F_1$ scores of different decoding strategies. }
     \label{tab:rouge}
\end{table*}

Table \ref{tab:rouge} shows the average ROUGE $F_1$ scores of all model combinations investigated. The attention-alignment mechanism promotes the quality of summaries by raising ROUGE-1 by $0.59$, ROUGE-2 by $0.4$, ROUGE-L by $1.16$ for PHT with beam search. Technically, the attention alignment mechanism could be applied to all hierarchical models with an attention mechanism. Further, Table \ref{tab:rouge} provides empirical evidence to Section \ref{why}, suggesting that extractive probabilities (extProb) are not as good protocols as the paragraph attention for the optimal attention distribution, given the marked decline in the ROUGE scores in comparison to the proposed attention-alignment mechanism.

Besides, the structural-coverage mechanism hinders the performance of MDS with reduced ROUGE scores. We print its beam-search scores and find consecutive zeros as the generated sequences get longer, resulted from the $t_{th}$ word attention on the $i_{th}$ paragraph ($\alpha_{t,i}$) remains lower than the its cumulative attention ($\sum_{t'<t} \alpha_{t', i}$). Therefore, it is concluded that the structural coverage regularization is not particularly suitable for word-by-word summarization with lengthy inputs, that come along with multiple documents.

\subsection{Human Evaluation}

To provide a better comparison between the MDS models, we select 4 representative summarization models with the best ROUGE performances in the human evaluation, namely T-DMCA \& Transformer-XL (flat structure), and Liu's HT \& PHT (hierarchical structure), and two decoding strategies including the original Beam search and attention alignment.

In the survey, multi-document summaries are scored from four perspectives, including (A) \textbf{Informativeness} (Does the summary include important information in the gold summary), (B) \textbf{Fluency} (Is the summary fluent and grammatically-correct), (C) \textbf{Conciseness} (Does the summary avoid repetition and redundancy), (D) \textbf{Factual consistency} (Does the summary avoid common sense mistakes such as wrong date, wrong location, or anything else against facts). We specify five ratings from \textit{Very poor} ($1$) to \textit{Very good} ($5$) to assess criteria (A)-(C), and three ratings of \textit{Much better} ($2$), \textit{Better} ($1$), and \textit{Hard to score} ($0$) to assess criterion (D). Twenty examples are randomly selected from generated summaries. Fifteen human evaluators participated in the experiment. 

The results are displayed in Table \ref{tab:hum}. The general observation is twofold. A) Discrepancy exists in the ROUGE and human evaluations. For instance, T-DMCA tends to yield higher human scores relative to Transformer-XL although ROUGE suggests the opposite. Given the merits and weaknesses of the different metrics, we focus on discussing results that exhibit consistency in different parts of evaluation. B) Given the lowest average mark, factual consistency appears to be the bottleneck of abstractive summarization models that hinders human experience on the machine generated summaries.

As far as the summarization models are concerned, PHT achieves the highest human evaluation scores in all four areas investigated. On the other hand, the other hierarchical baseline, Liu's HT, turns to be less competitive than the flat structures in terms of informativeness, conciseness and factual consistency, possibly due to its length limit of input. With regards to the optimal attention distribution of summaries, attention-alignment is proven effective in improving the hierarchical model.

\begin{table}[htbp]
   
    \label{tab:hum}
    \centering
    \begin{tabular}{lcccc}
    \toprule
     Model &  Informativeness & Fluency & Conciseness& Factual consistency \\
     \midrule
     T-DMCA & 3.69 & 3.66 & 3.82 & 3.04 \\
     Transformer-XL & 3.57 & 3.71 & 3.77 & 2.88 \\
     \midrule
     Liu's HT & 3.34 & 3.76 & 3.75 & 2.82 \\
     PHT & 4.11 & 3.97 & 3.81& 3.11\\  
     PHT with attAlign & \textbf{4.39} & \textbf{4.22} & \textbf{3.95} & \textbf{3.35}\\
     \midrule
     Average & 3.77 &3.89& 3.84 & 3.05\\
     \bottomrule
    \end{tabular}
    \caption{Human evaluation results.}
\end{table}

\section{Analysis}
This section discusses the experimental results obtained from Wikisum using different baseline models. Through the preliminary analysis on PHT, we intend to obtain an initial understanding of the hierarchical model on its capacity to better express the cross-document relationship, as well as the associated computational cost. The improvements on the summary quality by PHT with attention alignment is then investigated through the ROUGE analysis and human evaluation.

\subsection{Preliminary Analysis on PHT}

\subsubsection{The cross-document relationship}
Cross-document relationships could be reflected by the distribution of paragraph attentions. If a model assigns higher attention weights to more important paragraphs and vice versa, the model is believed to have greater capacity of capturing cross-document relationships. To analytically assess the models' performance in this aspect, we use paragraph attentions of reference summaries as the gold attention distribution, and its cosine similarity to the attention distribution of generated summaries as the evaluation metric. To model the paragraph attention of the reference, we compute the normalized tf-idf similarities between the gold summary and each input paragraph as the gold attention distribution. For the baseline models, the summation of token weights in each paragraph are computed to indicate each paragraph's attention, whilst PHT returns the paragraph attention distribution directly from its paragraph-level multi-head attention.

\begin{table}[htbp]
    \centering
    \begin{tabular}{lc}
    \toprule
     Model & Cosine similarity\\
     \midrule
     Lead & 0.8098\\
     \midrule

     Flat Transformer & 0.8143\\
     T-DMCA & 0.8654\\
     Transformer-XL & 0.8447\\
     \midrule
     Liu's HT & 0.8769\\
     PHT & \textbf{0.8936}\\
        \bottomrule
    
    \end{tabular}
     \caption{Average cosine similarities between attention distributions of generated summaries and the reference. }
   \label{tab:cap}
\end{table}

As suggested by Table \ref{tab:cap}, hierarchical structures place significant improvements on the flat models in learning cross-document dependencies by assigning paragraph attentions in a way that is closer to the gold summaries. Moreover, PHT generates summaries of the greatest similarity 89.36\% with the gold summaries, most likely due to its paragraph-level multi-head attention in addition to the token-level one, allowing the exchanging of cross-document information.

\subsubsection{Computational efficiency}
This section assesses the computational efficiency of PHT comparing to other neural abstractive models in three aspects, i.e. the memory usage, parameter size and validation speed. We uniformly hire the 3-layer architecture and 1600 input tokens in this part to ensure fairness. During the experiment, we increase the batch size until out of memory in a 2080ti GPU, and the model with the maximum batch size occupies the lowest memory space. To measure the parameter size, we count the number of parameters in the neural network. Finally, we run each trained model in the validation set (38,144 samples), and the average time consumed in each checkpoint is used to evaluate the speed of forward-propagating in the model. 

\begin{table}[htbp]
    \centering
    \begin{tabular}{lccc}
    \toprule
     Model & Max Batch Size & Parameters (MB) & Validation Speed (s)\\
     \midrule
     Flat Transformer & 11 & 165.0 & 634 \\
     T-DMCA & 10 & 131.1 & 656\\
     Transformer-XL & 8 & \textbf{130.4}& \textbf{489}\\
     \midrule
     Liu's HT & 11 & 190.8 & 639\\
     PHT & \textbf{17} & 182.4 & 601 \\
        \bottomrule
    \end{tabular}
        \caption{Computational efficiency.}
        \label{tab:eff}
\end{table}

As indicated by higher batch sizes in Table \ref{tab:eff}, models in the hierarchical structure (second panel) appears to be overall more memory-saving than those in the flat structure (first panel), with higher requirements on the parameters. In particular, models based on the Transformer-decoder, i.e. T-DMCA and Transformer-XL, demonstrate absolute superiority in reducing the parameter size. As for the speed of forward-propagating, Transformer-XL dominates due to its recurrent mechanism, whereas others share close performance in the inference speed. Between the two hierarchical models, PHT is proven to outperform Liu's HT in all three aspects, due to its parallel, rather than sequential, computation of the word \& paragraph-level attention mechanisms.

As an extension on attention alignment to explore its potential application on summary compression, we further argue that the algorithm provides an easy way to compress source paragraphs before inference -- by ranking paragraph attention weights according to their predicted values and choosing the top $s$ paragraphs with highest attentions. Table \ref{tab:compress} presents the results given different values of $s$. According to the ROUGE-Recall and ROUGE-$F_1$ scores, the tested compression mechanism improves original summaries by limiting the number of paragraphs to $[20:25]$. 

\begin{table}[htbp]
   \centering
    \begin{tabular}{lccccccccc}
    \toprule
      \multirow{2}*{$s$} & \multicolumn{3}{c}{ROUGE-1} & \multicolumn{3}{c}{ROUGE-2} & \multicolumn{3}{c}{ROUGE-L}  \\
        \cmidrule{2-10}
        & $F_1$ & $P$ & $R$ &   $F_1$ & $P$ & $R$   & $F_1$ & $P$ & $R$ \\
        \midrule
        5 & -2.1 & -3.3 & -1.2 & -2.0 & -2.5 & -1.8 & -2.1 & -3.8 & -1.5\\
        10 & -1.2 & -3.0 & -0.2 & -1.2 & -2.2 & -0.9 & -1.3 & -3.4 & -0.5\\
        15 & -0.4 & -2.6 & +\textbf{0.7} & -0.5 & -1.9 & 0 & -0.1 & -2.7 & +\textbf{0.5} \\
        \underline{20} & 0 & -1.5 & +\textbf{0.7} & -0.1 & -1.2 & +\textbf{0.2} & +\textbf{0.2} & -1.9 & +\textbf{0.5} \\
        \underline{25} & +\textbf{0.1} & -0.9 & +\textbf{0.6} & -0.1 & -0.8 & +\textbf{0.3} & +\textbf{0.1} & -1.0 & +\textbf{0.5} \\
        30 & 42.6 & 58.6 & 37.8 & 20.8 & 30.6 & 18.5 & 35.7 & 55.6 & 35.9\\
        \bottomrule
    \end{tabular}
     \caption{Average ROUGE scores with different $s$, No compression when $s=30$. $F_1$: $F_1$ score, $R$: Recall, $P$: Precision.}
    \label{tab:compress}
\end{table}

\section{Conclusion}\label{conclusion}

This study develops a Parallel Hierarchical Transformer with attention alignment inference for multi-document summarization. Using the Wikisum dataset, we empirically show that the proposed hierarchical architecture with token- and paragraph-level multi-head attentions excels in capturing the cross-document relationship of lengthy sources, and generates summaries of greater quality than other existing Transformer-based models. Further, the paragraph-level attention-alignment algorithm is designed to address the coverage issue by predicting the optimal attention distribution according to the multi-document sources. In theory, the decoding strategy has the potential to accommodate all seq2seq summarization models in the presence of the attention mechanism. Our experiment shows that attention alignment places significant improvements on the summaries generated by the original beam search. 

Given the fact that the attention mechanism is nowadays almost a necessity in the seq2seq architecture, the authors target at investigating the capacity of word-level attention-alignment in the future study. The application of word-level attention alignment is no longer confined to the hierarchical architecture and can be adopted to all attention-based seq2seq models including the pre-trained model BART \citep{lewis-etal-2020-bart}. Different from paragraph-level attention alignment, word-level attention alignment though requires the processing of numerous attention units, bringing challenges in obtaining the optimal attention distribution where a dynamic scoring function might be developed for text decoding. 

\section{Acknowledgements}
This work was supported by Xi'an Jiaotong-Liverpool University Key Programme Special Fund KSF-A-14.

\newpage
\bibliography{Main}
\end{document}